\pdfoutput=1
\documentclass[10pt,twocolumn,letterpaper]{article}

\usepackage{cvpr}              % To produce the CAMERA-READY version
\usepackage[linesnumbered,ruled,vlined]{algorithm2e}

\usepackage[dvipsnames]{xcolor}

\definecolor{cvprblue}{rgb}{0.21,0.49,0.74}
\usepackage[pagebackref,breaklinks,colorlinks,citecolor=cvprblue]{hyperref}
\usepackage{subcaption}
\usepackage{tcolorbox}
\usepackage{subcaption}
\usepackage[accsupp]{axessibility}

\SetKwInput{KwInput}{Input}

\SetCommentSty{mycommfont}

\newcommand{\methodstepA}{AtomLenz}
\newcommand{\methodstepAtitle}{ATOM-Level ENtity localiZer}
\newcommand{\methodstepB}{EditKT*}

\newcommand{\methodstepC}{ChemExpert}
\newcommand{\methodstepCtitle}{ChemExpert: Combination of experts}
\def\paperID{14628} % *** Enter the Paper ID here
\def\confName{CVPR}
\def\confYear{2024}

\title{Atom-Level Optical Chemical Structure Recognition with Limited Supervision}
\author{Martijn Oldenhof\textsuperscript{\rm 1}\hspace{1cm}
Edward De Brouwer\textsuperscript{\rm 1,\rm 2}\hspace{1cm} Adam Arany\textsuperscript{\rm 1}\hspace{1cm}Yves Moreau\textsuperscript{\rm 1}
\\
\textsuperscript{\rm 1}ESAT - STADIUS, KU Leuven, Belgium\\
\textsuperscript{\rm 2}Yale University, USA\\
{\tt\small \{martijn.oldenhof, edward.debrouwer, adam.arany, yves.moreau\}@esat.kuleuven.be}
}

\begin{document}
\newcommand{\edward}[1]{\textcolor{purple}{#1}}
\newcommand{\martijn}[1]{\textcolor{green}{#1}}
\maketitle
\begin{abstract}

Identifying the chemical structure from a graphical representation, or image, of a molecule is a challenging pattern recognition task that would greatly benefit drug development.
%The task of identifying chemical structures depicted in 2D images is a notable challenge in the machine learning field. 
Yet, existing methods for chemical structure recognition do not typically generalize well, and show diminished effectiveness when confronted with domains where data is sparse, or costly to generate, such as hand-drawn molecule images. 
%Despite the numerous existing methods, their effectiveness often diminishes when confronted with domains where
To address this limitation, we propose a new chemical structure recognition tool that delivers state-of-the-art performance and can adapt to new domains with a limited number of data samples and supervision. Unlike previous approaches, our method provides atom-level localization, and can therefore segment the image into the different atoms and bonds. Our model is the first model to perform OCSR with atom-level entity detection with only SMILES supervision.
%Our method operates by using a self-labeling strategy to generate atom-level annotations, enhancing the dataset's value, and using this enriched representation to fine-tune the model to a new domain.
Through rigorous and extensive benchmarking, we demonstrate the preeminence of our chemical 
structure recognition approach in terms of data efficiency, accuracy, and atom-level entity prediction.

%when enriched with atom-level entity information, showcasing unprecedented data efficiency and accuracy.
%explicitly crafted to excel in low-data domains, such as hand-drawn images, while concurrently providing atom-level localization.
%In this research, we present a pioneering chemical structure recognition tool explicitly crafted to excel in low-data domains, such as hand-drawn images, while concurrently providing atom-level localization. 
%To validate our tool's performance, we conduct a rigorous benchmark against state-of-the-art solutions within the field. 
%Remarkably, our tool not only competes with but frequently outperforms its counterparts, all while delivering atom-level localization—an attribute notably absent in most existing tools. 
%Our approach leverages a novel training strategy, initially utilizing a dataset of hand-drawn images paired with their corresponding SMILES representations (which encode molecular graph structure). We employ a self-labeling method to generate atom-level annotations, enhancing the dataset's value. 

\end{abstract}
\section{Introduction}
\label{sec:intro}

Molecules and chemical reactions represent the tokens of the language of chemistry, which underlies applications such as drug or new materials discovery. Molecules can be represented by a molecular formula (\emph{e.g.}, C$_8$H$_{10}$N$_4$O$_2$), or preferably by a more detailed structural formula—a graphical representation showcasing the spatial arrangement of atoms in the molecule. Isomers, molecules sharing the same molecular formulas but differing in spatial atom arrangement, typically exhibit distinct chemical and physical properties (as illustrated in Supplementary Material (SM) Section~\ref{sec:isomer}).
%arranged in three-dimensional space (\edward{Example ?})\footnote{ Different compounds may have same molecular representation, making it a poor representation. \edward{would be cool to have an example of two drugs with same formula and very different structures in the appendix (e.g. Cocaine and Sugar).}}.
Structural molecular formulas are thus ubiquitous in chemistry publications, lab notes, patents, or text books. This prevalence motivates the development of automatic pipelines to perform chemical structure recognition, parsing structural formulas from images. Such ability promises more efficient scientific literature browsing, automatic lab notes transcription, or chemical data mining, among others.

\begin{figure*}[htbp]
  \centering
   \includegraphics[width=0.8\linewidth]{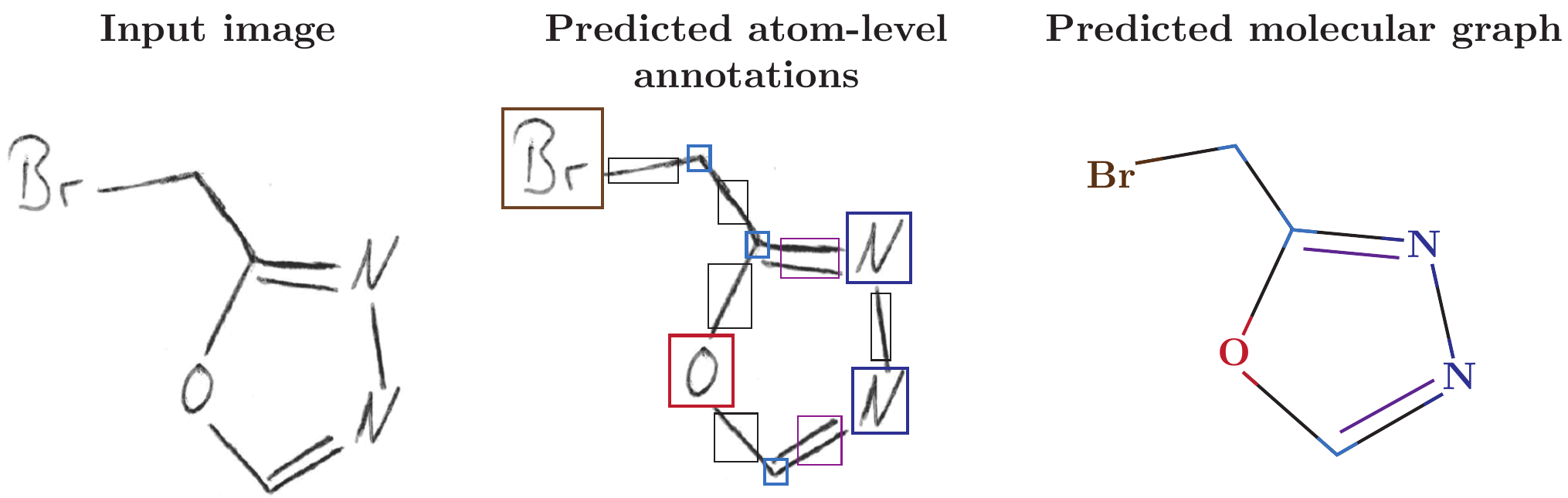}

   \caption{\textbf{Problem setup.} Our chemical structure recognition method takes an input image and predict atom-level entities predictions (atoms, bonds, charges, and stereocenters). This rich annotation can then be used to construct the molecular graph. Atom-level entity prediction models, like ours, predict both the atom-level annotations and molecular graph from the image. In contrast, molecular graph predictions models only predict the molecular graph and do not provide any localization in the original image.}
   \label{fig:idea}
\end{figure*}

Recent advances in computer vision have allowed the development of several chemical structure recognition tools~\cite{rajan2020decimer,clevert2021img2mol,oldenhof2020chemgrapher}.
These tools can be classified into molecular graph predictions methods and atom-level entity prediction methods. Molecular graph prediction methods only use limited image annotation, such as SMILES, a serial notation of a molecule~\cite{weininger1988smiles,rajan2020decimer,clevert2021img2mol}, and only predict the molecular graph. In contrast, atom-level entity prediction methods leverage richer image annotations for training the model, such as atom-level entity localization (\emph{i.e.}, individual atoms and bonds are annotated in the original image) \cite{qian2023molscribe,oldenhof2020chemgrapher}. These methods predict the molecular graph as well as the localization of the different components of the molecule in the original image. Figure~\ref{fig:idea} illustrates the different types of predictions for these two categories of models.

Previous research has shown that atom-level entity prediction methods typically enjoy better training sample efficiency, requiring less images for achieving the same level of performance~\cite{hormazabal2022cede}. This class of methods is also more interpretable. Atom-level entity annotation can indeed help identify the atoms that will be part of new chemical bond in a reaction, and can also facilitate human evaluation and correction when necessary, opening the way for synergistic human-in-the-loop training strategies~\cite{qian2023molscribe}. Nevertheless, these advantages are compensated by the necessity to provide rich image annotation in the training data. Unfortunately, such supervision is often unavailable in many data domains, such as hand-drawn images. Yet, hand-drawn images represent a prevalent format in chemical notations and sketches. The strict dependency of existing atom-level entity prediction methods on rich image annotation thus prevents their deployment to crucial data domains.

%In the field of drug discovery, the graph structure of a molecule serves as a fundamental input for machine learning pipelines.
%The accurate extraction of these intricate molecular graph structures from images is paramount for advancing drug development. This task falls under the purview of chemical structure recognition, a critical area within the domain of computer vision and machine learning.
%olecules and chemical reactions are the language of chemistry and drug discovery. As such, they have been given a dedicated writing system: molecular and structural formulas. A molecular formula is a representation of a molecule that uses chemical symbols to indicate the types of atoms followed by subscripts to show the number of atoms of each type in the molecule. For instance, on may write caffeine as
%\edward{different compounds may have same molecular representation. Chemists don't even think about using it as a representation. Only basic components would use the formula.}
%\chemfig{*6(([,0.5]=O)-N([6,0.7]-)-*5(-N=-N([:60, 0.7]-)-=)--([,0.5]=O)-N([:150, 0.7]-)-)}
%Molecular formulas are also used as abbreviations for the names of compounds.
%These promises have fueled the development of several chemical structure recognition tools~\cite{rajan2020decimer,clevert2021img2mol,oldenhof2020chemgrapher}.

Our research addresses these limitations by introducing a state-of-the-art chemical structure recognition tool, which (1) predicts a molecular graph from images, (2) provides atom-level localization in the original image, and (3) adapts to new data domain with a limited number of data samples and supervision. Our architecture relies on a object detection backbone coupled with a graph construction strategy that is pretrained on synthetic data, where the localization of each atom-level entity is known. We then leverage the atom-level entity localization, coupled with a efficient self-relabeling strategy, to aptly transfer to new domains where no localization is available (typically only image-SMILES pairs are available). This results in a state-of-the-art  and highly data efficient architecture, as demonstrated by our rigorous benchmarking.

\emph{Key contributions}: (1) We propose a novel framework for chemical structure recognition that predicts atom-level localizations trained on a target domain with only SMILES supervision. (2) We show that our method results in state-of-the-art performance on challenging hand-drawn molecule images, with a remarkable data efficiency. (3) We release a new curated dataset containing hand-drawn molecules with atom-level annotations.

%(1) We propose a new framework to enrich datasets and self-label atom-level annotations given images and SMILES. (2) We propose a innovative chemical recognition tool which can be trained in a data efficient way. (3) We contribute newly atom-level annotations for hand-drawn images in order to enable further research in this domain.

%\edward{Our training scheme can enhanced other architectures.}
%\edward{We are better in rare atoms because we are efficient sample-wise}
%\edward{we demonstrate experimentally that incorporating atom and edge localization data into the training process not only enhances result interpretation but also substantially contributes to the improved performance of the recognition model.}

%All contributions hold significant promise for advancing the landscape of chemical structure recognition, with far-reaching implications for drug discovery, chemistry, and related fields. In the spirit of open science, all datasets and code are readily available on GitHub, fostering collaboration and further innovation within the scientific community.

Our implementation is available on Github: \newline \url{https://github.com/molden/atomlenz}% (more info in SM Section~\ref{app:code}).

%\subsection{Miscellaneous}

%\begin{figure*}
%  \centering
%  \begin{subfigure}{0.68\linewidth}
%    \fbox{\rule{0pt}{2in} \rule{.9\linewidth}{0pt}}
%    \caption{An example of a subfigure.}
%    \label{fig:short-a}
%  \end{subfigure}
%  \hfill
%  \begin{subfigure}{0.28\linewidth}
%    \fbox{\rule{0pt}{2in} \rule{.9\linewidth}{0pt}}
%    \caption{Another example of a subfigure.}
%    \label{fig:short-b}
%  \end{subfigure}
%  \caption{Example of a short caption, which should be centered.}
%  \label{fig:short}
%\end{figure*}

\section{Background}
\label{sec:formatting}

Our work builds upon the chemical structure recognition literature and takes an object detection approach for solving this task. To enable fine-tuning of the model where no atom-level annotations are present, we leverage advances in weakly supervised object detection.

\subsection{Chemical structure recognition}

%\edward{probably focus more on the distinction between the different methods. we have discussed SMILES in the intro.}

%Paragraphs
%OCSR - SMILES
% Graphs
% Atom-level entity recognition

Optimal chemical structure recognition consists in inferring the structural formulae of a chemical compound based on an image representation of it. The large majority of  existing methods performing this task take the image as input, and predict the SMILES (simplified molecular-input line-entry system) representation of the molecule~\cite{weininger1988smiles}. SMILES consists of strings of ASCII characters that are obtained by printing the chemical symbols encountered in a depth-first tree traversal of the molecular graph. This serial notation provides, at first sight, a convenient representation for training machine learning models, while encoding geometric information about the molecular graph. This justified the popularity of SMILES-based chemical structure recognition models~\cite{rajan2020decimer,clevert2021img2mol}.

Nevertheless, SMILES do not provide a natural chemical representation and do not readily encode the geometric properties of the molecules. This hampers the trainability of the underlying machine learning model~\cite{hormazabal2022cede}. This limitation motivated the development of methods predicting the molecular graph and capable of leveraging richer image annotations, such as atom-level localizations~\cite{qian2023molscribe,oldenhof2020chemgrapher}.
Our work belongs to this category and therefore inherits these strengths. However, we extend previous approaches by providing a mechanism to fine-tune the model to new data domains where only SMILES annotations are available.

\subsection{Object detection}

Our architecture draws heavily on the literature on objection detection in images ~\cite{girshick2014rich, girshick2015fast,ren2015faster,redmon2016you,zhu2020deformable}, which underlies a wide array of high-level machine learning applications~\citep{behl2017bounding,giunchiglia2022road,tomei2019art2real,hameed2021content}. We refer to ~\cite{zou2023object} for a recent review of the field. Training object detection models typically requires comprehensive image annotations, such as the precise coordinates of bounding boxes and the associated labels for every object contained within each image. However, and crucially for our application, these annotations are not consistently accessible within certain domains of interest. This scarcity of detailed annotations has spurred the development of \emph{weakly supervised} object detection methods.

%The proposed method in this work relies heavily on object detection methods which constitutes a fundamental capability in a wide array of high-level machine learning applications~\citep{behl2017bounding,giunchiglia2022road,tomei2019art2real,hameed2021content}. Object detection is already a field which has evolved a lot in the last 20 years~\cite{zou2023object}. Recent object detection models can be two stage models~\cite{girshick2014rich, girshick2015fast,ren2015faster}, one stage model~\cite{redmon2016you} or even transformer based~\cite{zhu2020deformable}. 

%Achieving high-performance object detection models typically entails the provision of comprehensive image annotations, encompassing details like the precise coordinates of bounding boxes and the associated labels for every object contained within each image. 
%Instance-level annotations, crucial for object localization, are not consistently accessible within certain domains of interest. This scarcity of detailed annotations has spurred the development of weakly supervised object localization methods to address this challenge.

\subsection{Weakly supervised object detection}

This category of methods enables the training of object detection models without the necessity for precise bounding-box annotations. As a result, these approaches can be directly applied to target domains where such annotations are unavailable. Diverse variations of Weakly Supervised Object Detection (WSOD) architectures have emerged, relying on a range of implementations, including Multiple Instance Learning (MIL)~\cite{li2016weakly,song2014learning} and Class Activation Map (CAM)~\cite{zhou2016learning,bae2020rethinking} approaches. Other advanced WSOD techniques incorporate knowledge transfer from a source domain \cite{deselaers2012weakly,zhong2020boosting,uijlings2018revisiting,inoue2018cross,oldenhof2023weakly}. Among these approaches, ProbKT~\cite{oldenhof2023weakly} distinguishes itself as a versatile method, relying on probabilistic reasoning, which offers the capacity to train atom-level localization models using chemical background information obtained from SMILES and logical reasoning. Our architecture leverages this approach.

%Additionally, it is worth noting the method proposed by \citet{oldenhof2021self}, which facilitates self-labeling of atom-level annotations by aligning the true molecular graph with the predicted graph.
%For an in-depth exploration of WSOD methodologies, we recommend referring to Shao et al.'s comprehensive review [39].

%-------------------------------------------------------------------------

\section{Method}

% Overall description of the model.
% - object detection model
% - molecular graph construction
% - weakly supervised training.
% - relabeling
% - mixture of expert

\begin{figure*}[htbp]
  \centering
   \includegraphics[width=0.9\linewidth]{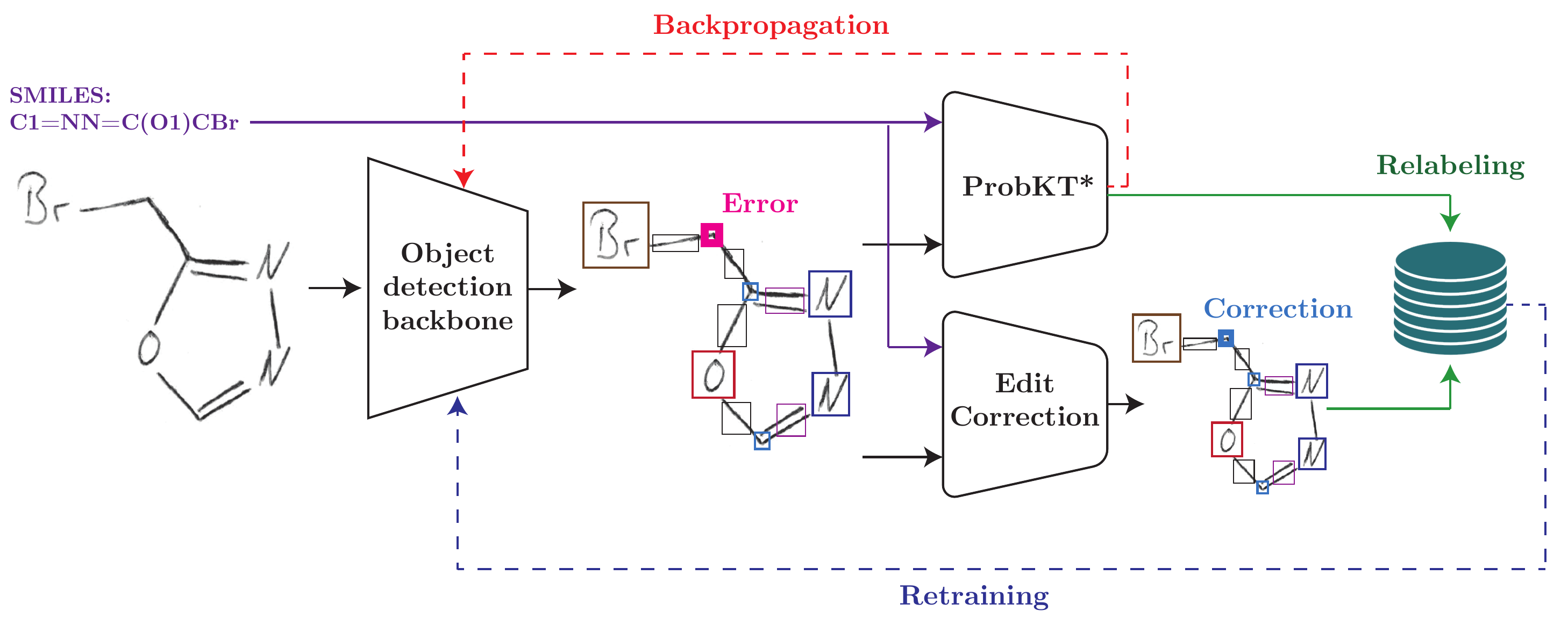}

   \caption{
   %The weakly supervised training scheme of the object detection backbone combining ProbKT* and Edit correction mechanism for pseudo-labeling data samples.
   \textbf{Weakly supervised training}. The weakly supervised training data set consists of images of molecular depictions paired with SMILES. The input image is used by the object detection backbone to predict the atom-level entities while the SMILES is used by ProbKT* and the edit-correction scheme. In the first phase, ProbKT* will perform backpropagation to update the object detection backbone using probabilistic reasoning. In the second phase, both ProbKT* and the edit-correction mechanism will generate pseudo-labels for the atom-level entities, which are used to retrain the object detection backbone.
   %\edward{update caption}
   }
   \label{fig:overall}
\end{figure*}

Our architecture is composed of four high-level modules: (1) an object detection backbone, which is trained on richly annotated images with atom-level entities, (2) a molecular graph constructor that assembles a molecular graph from the set of atom-level predictions, (3) a weakly supervised training scheme that enables fine-tuning the model on new domains without rich annotations. Additionally, we design a chemically informed combination of experts,~\methodstepC, that can further boost the prediction performance. The weakly supervised training scheme of the object detection backbone is visualized in Figure~\ref{fig:overall}.
%\edward{refer to the overall architecture figure}

%In this proposed approach, the model plays a central role in predicting the molecular graph complete with atom-level entity localization. This framework is further equipped with the capability to autonomously label atom-level entity annotations, enabling training on datasets containing both images and SMILES representations and no bounding boxes of atom-level entities. This section will delve deeper into the core model, followed by an examination of the self-labeling approach. Additionally, we introduce a Chem-Expert agent, which enhances performance by incorporating predictions from other chemical structure recognition tools through the evaluation and reasoning over the chemical validity of these predictions.

\subsection{Object detection backbone}

At the core of our architecture lies an object detection model that is responsible for detecting and labeling atom-level entities in the image. The objects in the image are therefore the atom-level entities such as atoms or bonds. While many object detection methods exist and can be used interchangeably in our architecture, we used the Faster RCNN model~\cite{ren2015faster} in our experiments. It is fast to train, robust, and simultaneously localizes and classifies all objects in a single step. The object detection backbone is trained by minimizing a multi-task loss $\mathcal{L}$ mixing a multi-class log loss $\mathcal{L}_{cls}$ applied on the predicted class $\hat{c}$ of the objects and a regression loss $\mathcal{L}_{reg}$ applied to the predicted bounding box coordinates $\hat{b}=(\hat{b}_x,\hat{b}_y,\hat{b}_w,\hat{b}_h)$ of the objects.

\begin{equation}
    \mathcal{L} = \mathcal{L}_{cls}(c,\hat{c}) + \mathcal{L}_{reg}(b,\hat{b})
\end{equation}

%The core model %, drawing inspiration from ChemGrapher~\cite{oldenhof2020chemgrapher},
%initially focuses on pinpointing various atom-level entities. Subsequently, it employs a geometrically informed graph construction algorithm to assemble the resulting molecular graph. To achieve atom-level localizations, an object detection model, Faster RCNN~\cite{ren2015faster}, is employed. This approach streamlines the process by simultaneously localizing and classifying all objects in a single step.
%, distinguishing itself from Chemgrapher, which utilizes a two-step approach, thereby simplifying the training process.
%The employed object detection model minimizes a multi-task loss $\mathcal{L}$ comprising two components: a multi-class log loss $\mathcal{L}_{cls}$ applied on the predicted class $\hat{c}$ of the objects and a regression loss $\mathcal{L}_{reg}$ applied to the predicted bounding box values $\hat{b}=(\hat{b}_x,\hat{b}_y,\hat{b}_w,\hat{b}_h)$ of the objects.

%\begin{equation}
%%    \mathcal{L} = \mathcal{L}_{cls}(c,\hat{c}) + \mathcal{L}_{reg}(b,\hat{b})
%\end{equation}

Bonds, atoms, or charges are graphically very different. To account for this heterogeneity, we train four distinct object detection models, each tailored to a specific atom-level entity: atoms ($\mathbf{O}^a$), bonds ($\mathbf{O}^b$), charge objects ($\mathbf{O}^c$), and  stereocenters ($\mathbf{O}^s$). A stereocenter, also known as a stereogenic center, refers to an atom within a molecule that carries groups in a way that exchanging any two of these groups results in a stereoisomer~\cite{mislow1984stereoisomerism}. More details about stereochemistry can be found in SM Section~\ref{sec:isomer}.

Each type of atom-level entity comprises multiple classes $c$ that the object detection backbone aims at labeling (via $\mathcal{L}_{cls}$). For instance, the bond object encompasses categories such as single bond, double bond, triple bond, aromatic bond, dashed bond, and wedged bond. Illustrations of different atom-level entity types and classes can be found in SM Section~\ref{sec:examples_detection}.

\subsection{Molecular graph constructor}

The output of the object detection backbone is a list of detected atom-level entities in the image, along with their predicted label and position. The objective of the molecular graph constructor is to assemble a chemically sound molecular graph from this list of predictions. This graph can then be easily converted to a SMILES format.

%Once all objects are predicted a geometric and chemical informed graph construction algorithm is applied to build the resulting molecular graph which can be converted to any format like a SMILES format.

%We refer to the atom predictions (both localization and labels) as $\mathbf{O}^a$, the bonds predictions as $\mathbf{O}^b$, the charge predictions as $\mathbf{O}^c$, and the stereocenters predictions as $\mathbf{O}^s$.
The output graph $G(V,E)$ is composed of a set of vertices $V$, corresponding to atoms, and edges $E$ corresponding to bonds. Each vertex and edge has a label (\emph{e.g.}, a node can be a carbon atom with a positive charge, an edge can be a double bond). Algorithm~\ref{alg:alg-1} outlines the series of steps involved in constructing the molecular graph from atom-level entity predictions $\mathbf{O}^a$ (atoms), $\mathbf{O}^b$ (bonds), $\mathbf{O}^c$ (charges), and $\mathbf{O}^s$ (stereocenters). It proceeds in four steps: (1) a filtering step, (2) a node creation step, (3) an edge creation step, and (4) a validation step.

The \textbf{filtering step} filters atoms from $\mathbf{O}^a$ that are severely overlapping on the image. When multiple atom objects show an Intersection over Union (IoU) score exceeding a specified threshold, only the object with the highest score is retained.

In the \textbf{node creation step}, we first attach charges to atom objects. Overlapping atom and charge objects exceeding a specific IoU threshold are then merged. The function $\mathrm{checkCharges}$ is responsible for determining which atom objects should carry a charge. A similar procedure is subsequently applied to identify atoms functioning as stereocenters, utilizing $\mathrm{checkStereoChem}$ for this purpose. The list of all atom objects, along with their potentially assigned charges or stereocenters are then added to the list of graph vertices.

In the \textbf{edge creation step}, we iterate over all bond objects and evaluate which vertices (atoms) overlap with these bonds, with the function $\mathrm{checkEdge}$. If only two candidate atoms are identified, the algorithm proceeds to add the edge to the graph. However, when more than two overlapping candidates emerge, the algorithm selects the two most probable ones, factoring in the orientation of the edge and the atoms involved.

Lastly, the \textbf{validation step} identifies potential chemistry-related issues through $\mathrm{ChemistryProblems}$ and endeavors to resolve them via $\mathrm{SolveChemistryProblems}$ to ensure the prediction of a chemically valid molecular graph. For instance, each chemical element is assigned a valence number, indicating the atom's capability to establish bonds with other atoms. If our algorithm detects within the output graph containing atoms with more bonds than their valence numbers permit, the $\mathrm{SolveChemistryProblems}$ function would attempt to remove bonds iteratively until a graph is formed without valence errors.

To maintain conciseness in our experiments and results, we refer to the combination of the object detection backbone and the molecular graph constructor as \methodstepA~(\methodstepAtitle). Further information regarding the subroutines used in the molecular graph constructor is available in SM Section~\ref{sec:subroutines}.

\begin{algorithm}
\SetAlgoLined
\KwInput{Atom-level predictions $\mathbf{O}^a$,$\mathbf{O}^b$,$\mathbf{O}^c$,$\mathbf{O}^s$}
\KwResult{Graph $G(V,E)$ , vertices $V$ and edges $E$}
% = $\mathrm{PredictObjects}(\mathbf{x})$

$\mathbf{\tilde{O}}^a$ = $\mathrm{filterAtoms}(\mathbf{O}^a)$  \tcp*{filtering step}

$V=$ [] \tcp*{vertices of graph}

\tcc{node creation step }

\For{$\mathbf{o}^a$  \textbf{in} $\mathbf{\tilde{O}}^a$}{
  $\mathbf{o}_c^a$ = $\mathrm{checkCharges}(\mathbf{O}^c, \mathbf{o}^a)$

  $\mathbf{o}_{c,s}^a$ = $\mathrm{checkStereoChem}(\mathbf{O}^s, \mathbf{o}_{c}^a)$

  $V.\mathrm{appendAtom}(\mathbf{o}_{c,s}^a)$

 }

 $E=$ [] \tcp*{edges of graph}

\tcc{edge creation step}

 \For{$\mathbf{o}^b$  \textbf{in} $\mathbf{O}^b$}{
 $\mathrm{candAtoms} = \mathrm{checkEdge}(V, \mathbf{o}^b)$

 \If{$\mathrm{len}(\mathrm{candAtoms})==2$}{
 $E.\mathrm{appendBond}(\mathbf{o}^b, \mathrm{candAtoms})$

 }
 \If{$\mathrm{len}(\mathrm{candAtoms})>2$}{
 $\mathrm{filteredAtoms} = \mathrm{filterCands}(\mathrm{candAtoms})$
 $E.\mathrm{appendBond}(\mathbf{o}^b, \mathrm{filteredAtoms})$
 }
 }

\tcc{validation step}

 \If{$\mathrm{ChemistryProblems}(G(V,E))==\mathrm{True}$}{
 $G(V,E)=\mathrm{SolveChemistryProblems}(G(V,E))$
 }
 \caption{Molecular graph constructor}
 \label{alg:alg-1}
\end{algorithm}

\subsection{Weakly supervised training}

Our architecture uses an object detection backbone to predict atom-level entities, which requires rich image annotations, such bounding boxes for every object type, including atoms, bonds, charges, and stereocenters within the images. While such a level of supervision can be obtained synthetically with tools like RDKit~\cite{rdkit}, it is usually not available in real-world target domains, such as hand-drawn images.
In such domains, only SMILES are typically available. To enable the fine-tuning of the object detection backbone with only SMILES information, we use a weakly supervised training mechanism that combines (1) a probabilistic logical reasoning module that allows to differentiate through the object detection backbone with only weak supervision, and (2) a graph edit-correction mechanism that allows fine-tuning on less frequent atoms and bonds. A graphical outline of the weakly supervised training procedure is given in Figure~\ref{fig:overall}.

%This limits atom-level predictions methods to generalize to new domains. To address this limitation, we propose a weakly supervised training mechanism that allows the fine-tuning of the object detection backbone when only SMILES labels are present. Our weakly supervised training consists of (1) a probabilistic logical reasoning module that allows to differentiate through the object detection backbone with only weak supervision, (2) a image relabeling step that enrich the target dataset with atom-level entity annotations, enabling direct fine-tuning of the object detection backbone, and (3) a graph edit-correction mechanism that allows fine-tuning on less frequent atoms and bonds.

%Given that our core model relies on object detection for training, it needs bounding boxes for every object type, including atoms, bonds, charges, and stereocenters within the images.

%While generating such bounding boxes synthetically is feasible using tools like RDKit~\cite{rdkit}, it becomes challenging in target domains such as hand-drawn images. Therefore, our method encompasses a self-labeling mechanism designed to automatically generate these bounding boxes when both images and the corresponding SMILES representations of the chemical structure are available.

\subsubsection*{Backpropagation with weak supervision}

To update the weights of the object detection backbone with only SMILES supervision, we use the ProbKT~\cite{oldenhof2023weakly} framework. This weakly supervised domain adaption technique uses probabilistic programming for fine-tuning object detection models with a wide range of supervision signals, and is thus particularly suited for our application. In our experiments, we used ProbKT$^*$, a computationally efficient variant of ProbKT that relies on Hungarian matching.

ProbKT$^*$ allows differentiating through the object detection backbone with only SMILES supervision. For better performance, it also includes a relabeling mechanism, where confident predictions are used as new atom-level annotation of the target domain images. This strategy effectively creates a richly annotated dataset that can be used to fine-tune the object detection backbone directly.

\subsubsection*{Edit-correction mechanism}

While ProbKT$^*$ is generally effective at performing weakly supervised domain adaptation, it fails when dealing with rare atoms or bonds types. We therefore combine ProbKT$^*$ with a new edit-correction mechanism~\cite{oldenhof2021self} designed to detect and rectify minor errors in model predictions.
%This edit-correction mechanism utilizes these corrections to annotate the image with atom-level predictions.
Based on the SMILES, one can generate a reference true graph, although not aligned on the original image. The edit-correction mechanism solves an optimization problem that aims at finding the smallest edit on the predicted graph such that the true and corrected graphs are isomorphic. While this optimization would be intractable in general, focusing on small edits makes it computationally feasible. If such a correction is found, it is used to annotate the image which can then be used to fine-tune the object detection backbone.

\subsubsection*{Combined weakly supervised training}

When fine-tuning on a new target domain, we proceed by iteratively applying ProbKT$^*$ and the edit-correction scheme. In practice, we start with a few iterations of ProbKT$^*$. We then use multiple iterations of the edit-correction scheme until the validation performance stops improving. For sake of conciseness in our experiments results, we abbreviate the combination of both approaches as \methodstepB.

%Both of these approaches, as described in~\cite{oldenhof2021self} and~\cite{oldenhof2023weakly}, follow iterative processes. In practical implementation, the initial step typically involves applying ProbKT$^*$ for one or more iterations. Following this, the method based on graph edits, as presented in~\cite{oldenhof2021self}, is applied, also potentially undergoing multiple iterations until convergence is achieved. We abbreviate the combination of both approaches discussed in ~\cite{oldenhof2021self} and~\cite{oldenhof2023weakly} as \methodstepB ~for simplicity.

\subsection{\methodstepCtitle}

For the final prediction of our architecture, we propose to use a combination of experts, which is constrained by chemical soundness of the model predictions. We call this module \methodstepC. It relies on a list of chemical structure recognition tools, ordered by the user's preference in terms of the predictions. The first tool serves as the most trusted model. At inference time, \methodstepC~iteratively checks the validity of the prediction of each model in the list. If a chemical issue is identified, the agent evaluates the next model in the list. The module returns the prediction of the first model with no chemical issues detected. This strategy  enables us to incorporate predictions from additional tools alongside those generated by our core model, thereby improving predictive performance. In practice, we use a combination of DECIMER~\cite{decimerai} and our approach.

%This agent examines the SMILES predictions generated by these tools and assesses the chemical validity of these predictions.

%It is configured by establishing an ordered list of tools, with the first tool in the list serving as the default choice for trust. If any chemical issues are identified in the predictions made by the default tool, the agent proceeds to evaluate the predictions of the next tool in the list until no further chemical problems are detected or the last tool in the list is reached. The resulting prediction is then designated as the final output. The order of the list is tuned based on validation data. The agent enables us to incorporate predictions from additional tools alongside those generated by our core model, thereby improving predictive performance.

%\edward{in the experiments we use DECIMER + ours}

%\colorbox{BurntOrange}{TODO describe how self-labeling \cite{oldenhof2021self} }
%\colorbox{BurntOrange}{and ProbKT \cite{oldenhof2023weakly} are combined}

\section{Datasets}\label{sec:datasets}

\subsection{Synthetically generated dataset}
For the pretraining of the object detection models, we generate images synthetically using RdKit~\cite{rdkit} and Indigo~\cite{pavlov2011indigo} paired with bounding boxes delineating all objects within, including atoms, bonds, charges, and stereocenters, similarly to what is used in other chemical structure recognition tools~\cite{oldenhof2020chemgrapher,qian2023molscribe}. Specifically, we collect approximately 214,000 chemical compounds in SMILES format from the ChEMBL~\cite{gaulton2017chembl} database. To enhance the method's resilience to stylistic variations, we introduce variability in elements such as fonts, font sizes, line widths, and the spacing between multiple bonds during image generation. More details on this dataset can be found in SM Section~\ref{app:code}.

\subsection{Hand-drawn images datasets}

To facilitate the training, fine-tuning, and testing of our models on hand-drawn images, we meticulously curate multiple datasets. We begin with the dataset introduced by \citet{handdrawndataset}, which consists of hand-drawn chemical depictions matched with their corresponding SMILES representations. This dataset is partitioned into 4,070 samples for training and validation purposes, along with an additional 1,018 samples for testing. These sets are referred to as the \emph{hand-drawn training set} and the \emph{hand-drawn test set}.

\begin{figure*}
%\begin{tcolorbox}[title={Illustrations datasets samples},standard jigsaw,opacityback=0,colbacktitle=white,coltitle=black,]
\begin{subfigure}{0.3\textwidth}
\centering
\includegraphics[scale=0.1]{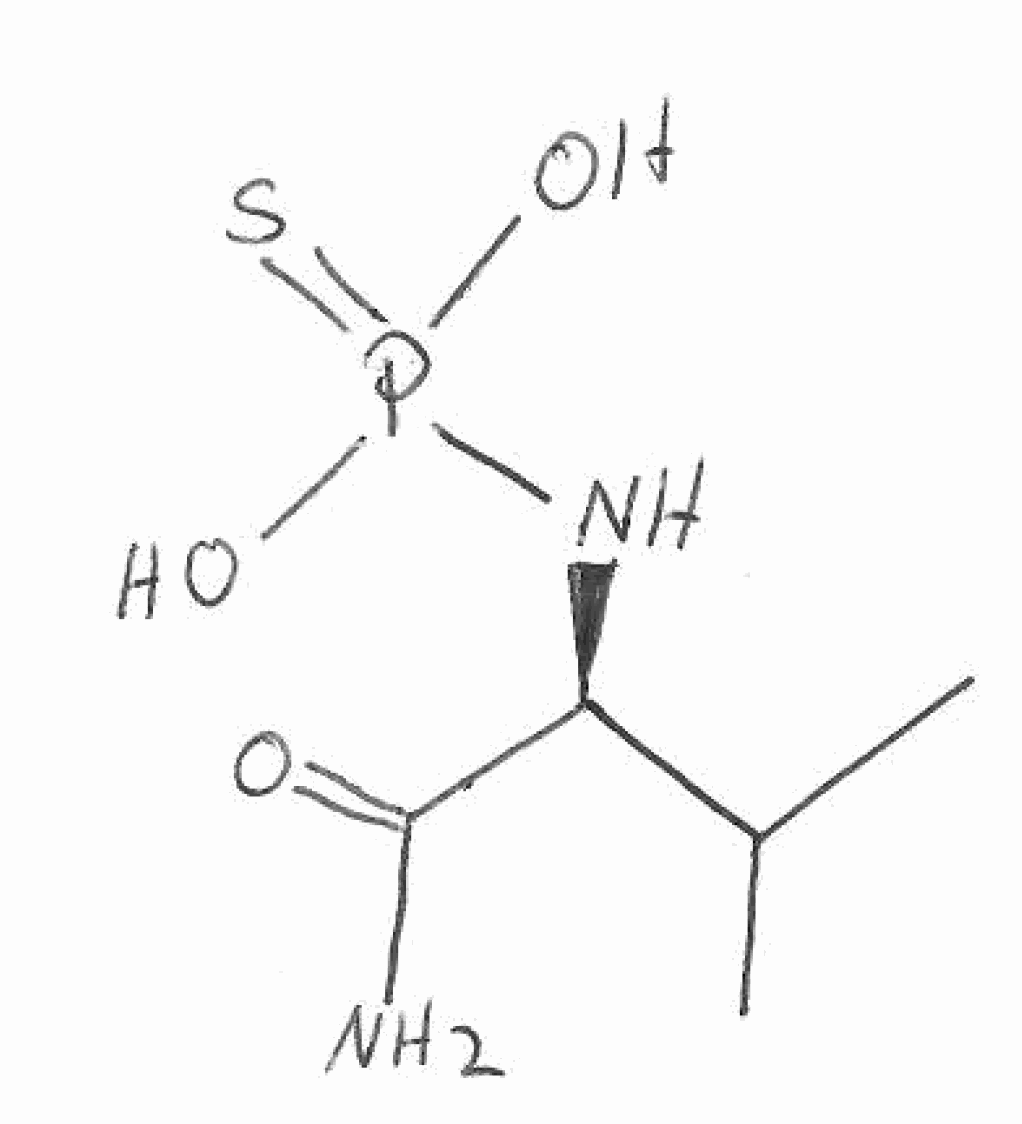}
\caption{Hand-drawn test set}
\label{fig:style_sub1}
\end{subfigure}
\begin{subfigure}{0.3\textwidth}
\centering
\includegraphics[scale=0.4]{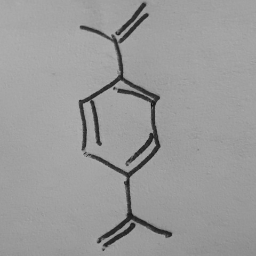}
\caption{ChemPix test set}
\label{fig:style_sub3}
\end{subfigure}
\begin{subfigure}{0.3\textwidth}
\centering
\includegraphics[scale=0.38]{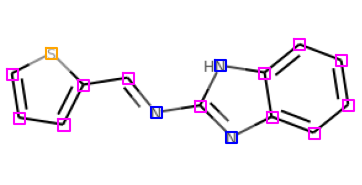}
\caption{Atom Localization test set, labeled with bounding boxes, here illustrated with different color for every atom type.}
\label{fig:style_sub4}
\end{subfigure}
%\end{tcolorbox}
\caption{\textbf{Image samples from the dataset.} Different example samples for the different datasets used in experiments. The hand-drawn and ChemPix datasets are used to assess the domain adaptation and out-of-domain performance. The atom localization dataset is used for testing object localization.}
\label{fig:example_samples}
\end{figure*}

%\begin{figure}[t]
%  \centering

%   \includegraphics[width=1\linewidth]{figures/36.png}

%   \caption{Example sample of target hand-drawn dataset.}
 %  \label{fig:handdrawn1}
%\end{figure}

%\begin{figure}[t]
%  \centering

%   \includegraphics[width=1\linewidth]{figures/45.png}

%   \caption{Example sample of target hand-drawn dataset.}
%   \label{fig:handdrawn2}
%\end{figure}

In addition to this primary test set, we incorporate an extra test dataset of 614 hand-drawn chemical depictions sourced from \citet{chempix}, which we call the \emph{ChemPix test set}, to further evaluate the performance of the models on hand-drawn images.

%\begin{figure}[t]
%  \centering

%   \includegraphics[width=1\linewidth]{author-kit-CVPR2024-v2/figures/ring_chempix.png}

%   \caption{Example sample of chempix hand-drawn dataset. Only carbon atoms are present in images.}
%   \label{fig:chempix1}
%\end{figure}

%\begin{figure}[t]
%  \centering

%   \includegraphics[width=1\linewidth]{author-kit-CVPR2024-v2/figures/triple_chempix.png}

%   \caption{Example sample of chempix hand-drawn dataset. Only carbon atoms are present in images.}
%   \label{fig:chempix2}
%\end{figure}

\subsection{Atom localization dataset}
To assess our models' capability for object localization, we employ a synthetically generated dataset using RdKit~\cite{rdkit} provided by \citet{oldenhof2023weakly}. This dataset encompasses 1,000 images depicting chemical structures, each meticulously annotated with bounding boxes outlining the positions and corresponding classes of all the atoms within the molecules. Example images for each test dataset are shown in Figure~\ref{fig:example_samples} and SM Section~\ref{app:code}.
%\colorbox{BurntOrange}{TODO Object (atom) detection test set ....}

%\begin{figure}[t]
%  \centering

%   \includegraphics[width=1\linewidth]{author-kit-CVPR2024-v2/figures/25.png}

%   \caption{Example sample of atom localization dataset. These images are synthetically generated images.\martijn{TODO: add bounding boxes around atoms}}
%   \label{fig:chempix2}
%\end{figure}

\section{Experiments and Results}

%\edward{Take home : - we are still state of the art on the training domain (Table 1 - hand-drawn images) - target domain (Table 2 - hand-drawn images - fine-tune on data from Table 1 and evaluated on new domain) - Atom level localization (Table 3) - data efficiency (Table 4). We should add figures about the datasets. \\
%Figure with the correlation between the frequency of occurrence of atoms and the atom-level prediction accuracy.}

Our experiments investigate the performance of our approach on hand-drawn images of chemical structures, as this domain suffers from limited data availability and has been shown to be a weak point of existing tools. We evaluate our architecture and state-of-the-art baselines on four distinct fronts: (1) molecular recognition on the new target domain (\emph{i.e.,} only predicting the SMILES), (2) atom-level entity localization, (3) training efficiency (when retrained from scratch), and (4) model evaluation per atom and bond type.

\paragraph{Baselines} We compare our architecture with the following baselines.
\textbf{DECIMER}~\cite{rajan2020decimer} is an image-transformer approach trained on more than 400 million synthetically generated data samples. The authors of DECIMER have also introduced a version  specifically tailored for hand-drawn images (DECIMER fine-tuned~\cite{decimerai}). Although it is trained on synthetically generated images, the training dataset of this version mimics the style of hand-drawn images more closely.
\textbf{Img2Mol}~\cite{clevert2021img2mol} integrates a deep convolutional neural network trained on molecule depictions (11 million synthetically generated images) with a pretrained decoder.
\textbf{MolScribe}~\cite{qian2023molscribe} and \textbf{ChemGrapher}~\cite{oldenhof2020chemgrapher} employ atom-level entity localization annotations in their training process on synthetically generated images. These are the only baselines that can predict atom-level annotations, alongside the SMILES predictions. ChemGrapher is trained on 114,000 generated images and MolScribe on 1 million generated images.
Lastly, \textbf{OSRA}~\cite{filippov2009optical} is a non-trainable, rule-based approach.

\begin{table*}[htbp]
  \centering
  \begin{tabular}{@{}lllll@{}}
    \toprule
    & \multicolumn{2}{c}{hand-drawn test set}  & \multicolumn{2}{c}{ChemPix test set} \\
    \midrule
    Method & Acc.($T=1$) & $\overline{T}$ & Acc.($T=1$) & $\overline{T}$ \\
    \midrule
    DECIMER (v2.2.0)\cite{rajan2020decimer} & 0.295 & 0.451 & 0.05 & 0.1 \\
    DECIMER fine-tuned(v2.2.0)\cite{decimerai} & 0.622 & 0.727 & 0.508 & 0.643\\
    Img2Mol\cite{clevert2021img2mol} & 0.084 & 0.275 & 0.015 & 0.084\\
    MolScribe\cite{qian2023molscribe} & 0.102 &  0.288  & 0.269 &  0.417\\
    ChemGrapher\cite{oldenhof2020chemgrapher} & 0.002 &  0.065 & 0.187 &  0.286\\
    OSRA\cite{filippov2009optical} & 0.006 &  0.065 & 0.047 &  0.071\\
    \midrule
    \methodstepA & 0.009 &  0.087 & 0.054 &  0.064\\
    \methodstepA+\methodstepB & 0.338 & 0.484 & 0.484 & 0.605\\
    \methodstepC(\cite{decimerai},\methodstepA+\methodstepB) &  \textbf{0.635} &  \textbf{0.749} &  \textbf{0.518} & \textbf{0.655}\\
    \bottomrule
  \end{tabular}
  \caption{Benchmark results on target domain (hand-drawn images test set) and out of domain ChemPix test set. Both the accuracy, computed by counting the instances where the predicted structures have identical structural ECFP6 descriptors (denoted by a Tanimoto ($T$) similarity of 1) and the average Tanimoto similarity ($\overline{T}$) are reported.
  %Accuracy is measured for all predictions with a Tanimotosimilarity ($T$) of 1 with true structural descriptor (ECFP6). Also the average Tanimoto similarity $\overline{T}$ is calculated for all predictions.  %\edward{best on target domain + chempix is OOD}
  }
  \label{tab:hand_drawn_combined}
\end{table*}

\subsection{Performance of molecular recognition on hand-drawn images}

We compare the performance of our approach with the baselines on the hand-drawn and ChemPix dataset. Results are given in Table~\ref{tab:hand_drawn_combined}. To assess the impact of our fine-tuning strategy, we evaluate three versions of our architecture. The first is a version trained on the synthetic dataset but not fine-tuned to the new hand-drawn dataset (\methodstepA). The second is fine-tuned to the hand-drawn dataset using EditKT$^*$. The third is~\methodstepC, combining DECIMER fine-tuned and~\methodstepA.  Performance of other combinations in~\methodstepC~ are reported in SM Section~\ref{sec:allresults}.

We assess the molecular structure prediction performance using accuracy and Tanimoto similarity. Tanimoto similarity ($T$)~\cite{tanimoto1958elementary}, a widely used metric for quantifying molecular similarity, to assess the resemblance between the model's predictions and the actual molecular graphs. Tanimoto similarity values range from 0 to 1, with higher values indicating greater similarity. A Tanimoto similarity of 1 indicates that the structural descriptors are identical or that they are matching `on-bits' in a binary fingerprint. The binary fingerprint employed to measure the Tanimoto similarity is the Extended-connectivity fingerprint~\cite{ecfp} with radius 3 (ECFP6) and fingerprint length of 2048.
%Crafted with precision to capture essential molecular features relevant to molecular activity, ECFPs (Extended-Connectivity Fingerprints)~\cite{ecfp} are generated through a customized adaptation of the Morgan~\cite{morgan} algorithm. This involves systematically traversing each atom in the molecule to extract all possible paths within a specified radius. Following this, every unique path undergoes hashing into a numerical value within a predetermined bit range. It is worth noting that the encoded fragment size expands proportionally with an increased radius.
More details on the calculation of the ECFP6 fingerprint and other fingerprints can be found in SM Section~\ref{sec:allresults}.

Our tables report both the accuracy, computed by counting the instances where the predicted structures have identical structural ECFP6 descriptors (denoted by a Tanimoto similarity of 1) and the average Tanimoto similarity. Additional measured metrics can be found in SM Section~\ref{sec:allresults}.

For both datasets \emph{hand-drawn test set} and \emph{ChemPix test set}, our \methodstepC~ performs best. This demonstrates that the combination of our approach with other baselines results in state-of-the-art performance. We further appreciate a significant increase in performance from EditKT$^*$, compared to the non-fine-tuned version, highlighting the effectiveness of our fine-tuning approach.

%In Table~\ref{tab:hand_drawn1}, we measure accuracy and average Tanimoto similarity on a hand-drawn image test set, a subset derived from a benchmark provided by ~\citet{handdrawndataset}. It is important to take into account that this test set belongs to the same domain as the training dataset used for refining our approach. Conversely, in Table~\ref{tab:hand_drawn2}, we evaluate the performance (accuracy and Tanimoto similarity) on a distinct test set obtained from \citet{chempix}, which also features hand-drawn images but differs in domain from the dataset of \citet{handdrawndataset}.

\subsection{Performance of atom-level localization}

In Table~\ref{tab:objects}, we assess the performance of the different methods in terms of their atom-level localization abilities. We employ a test set from \citet{oldenhof2023weakly}, which comprises images of chemical representations along with the corresponding atom objects. We use two evaluation metrics: the count accuracy (which can be evaluated without bounding box predictions), and the mean average precision (mAP) localization of the bounding boxes. The atoms count accuracy measures the ability to predict the correct number of atom types in each image. The average precision is computed as the weighted mean of precisions at various Intersect over Union (IoU) thresholds, with the weight reflecting the increase in recall from the previous threshold. Mean Average Precision represents the average of AP values across each class. We use the rather low IoU thresholds of $[0.05, 0.1, 0.15, 0.2, 0.25, 0.3, 0.35]$ in our experiments, considering the relatively small size of the bounding boxes of interest (see Figure~\ref{fig:style_sub4}), where significant overlap with the true bounding boxes is not anticipated. Methods that do not provide any form of localization are marked with n/a in the table.

In Table~\ref{tab:objects}, we observe the commendable localization performance of our pretrained backbone model (\methodstepA) and also note the high counting accuracy of DECIMER. We posit that both outcomes may be attributed to the nature of the test dataset, aligning with the characteristics of the images used for training both DECIMER and our core pretrained model (\methodstepA). Additionally, we note that DECIMER was trained on 2000 times more images than our architecture, and does not provide any localization.
%Also note the lower performance of our model after applying EditKT*. This model is retrained on significant lower number of samples compared to our pretrained model and uses mostly pseudo labels obtained from the hand-drawn target domain. Introducing more synthetically generated images into the mix could potentially enhance performance on this synthetically generated test set.
%\edward{it seems we should explain why fine-tuning hurts here....}
%Lastly, we assess the model's performance in object localization using the mAP (mean Average Precision) metric, along with the count accuracy of all objects present in the image.

\begin{table}[h]
  \centering
  \begin{tabular}{@{}lcc@{}}
    \toprule
    Method & Count Acc. & mAP \\
    \midrule
    DECIMER (v2.2.0)\cite{rajan2020decimer} & \textbf{0.973} & n/a \\
    DECIMER fine-tuned (v2.2.0)\cite{rajan2020decimer} & 0.97 & n/a \\
    Img2Mol\cite{clevert2021img2mol} & 0.929 & n/a  \\
    MolScribe\cite{qian2023molscribe} & 0.829 &  0.008 \\
    ChemGrapher\cite{oldenhof2020chemgrapher} & 0.248 &  0.002 \\
    OSRA\cite{filippov2009optical} & 0.255 &  n/a \\
    \midrule
    \methodstepA & 0.602 &  \textbf{0.801} \\
    %Ours+\methodstepB & 0.354 & 0.11 \\
    \bottomrule
  \end{tabular}
  \caption{Benchmark results on object (atom) detection test set to compare localization performance.
  %\edward{atom localization performance }
  }
  \label{tab:objects}
\end{table}

\subsection{Training efficiency}\label{training_eff}

Baseline architectures were trained on significantly larger number of images than our model. Molscribe uses 4 times more samples, while DECIMER uses a staggering 2000 times more. In Table~\ref{tab:training}, we evaluate the sample complexity of the different methods by retraining them from scratch on the same small training dataset, which mimics limited data availability scenarios.  We use the \emph{hand-drawn training set} (4,070 data samples), enriched with atom-level entity localization annotations generated using \methodstepB ~as the training dataset. We observe that the methods that leverage these atom-level entity annotations tend to fare better (ChemGrapher~\cite{oldenhof2020chemgrapher} and MolScribe~\cite{qian2023molscribe}) than the ones using SMILES as only supervision signal (DECIMER~\cite{rajan2020decimer} and Img2Mol~\cite{clevert2021img2mol}). Our approach significantly outperforms all baselines at this task, highlighting the remarkable data efficiency of our architecture. These findings align with those in the work of \citet{hormazabalcede}, where the author concluded that the use of atom-level entity annotations can enhance data efficiency during training. The hand-drawn images utilized in this experiment, along with the corresponding bounding box labels for 1417 images, we release as a novel annotated dataset. More info in the SM Section~\ref{app:code}.
%Some methods, such as ChemGrapher~\cite{oldenhof2020chemgrapher}, MolScribe~\cite{qian2023molscribe}, and our method (\methodstepA), have the capacity to leverage these atom-level entity annotations. In contrast, others, like DECIMER~\cite{rajan2020decimer} and Img2Mol~\cite{clevert2021img2mol}, do not use these annotations.

%This experiment aims to evaluate the efficacy of learning when presented with a limited dataset, a scenario commonly encountered in domains with limited data availability.

%All methods were trained using default hyperparameters, and while we acknowledge the potential for slight improvements through hyperparameter tuning, the noticeable performance gap between our approach and others underscores the remarkable data efficiency of our method. Our findings align with those in the work of \citet{hormazabalcede}, where the author concluded that the use of atom-level entity annotations can enhance data efficiency during training.

\begin{table}[h]
  \centering
  \begin{tabular}{@{}lcc@{}}
    \toprule
    Method & Acc.($T=1$) & $\overline{T}$ \\
    \midrule
    DECIMER (v2.2.0)\cite{rajan2020decimer} & 0.001 & 0.039  \\
    Img2Mol\cite{clevert2021img2mol} & 0.0 & 0.0867  \\
    MolScribe\cite{qian2023molscribe} & 0.013 &  0.0865 \\
    ChemGrapher\cite{oldenhof2020chemgrapher} & 0.004 & 0.067 \\
%    OSRA\cite{filippov2009optical} & - &  - \\
    \midrule
    \methodstepA & \textbf{0.338} &  \textbf{0.484} \\
    \bottomrule
  \end{tabular}
  \caption{All methods are retrained from scratch on same training dataset (4070 samples of hand-drawn images) to asses data efficiency. Benchmark results on hand-drawn images test set. Both the accuracy, computed by counting the instances where the predicted structures have identical structural ECFP6 descriptors (denoted by a Tanimoto ($T$) similarity of 1) and the average Tanimoto similarity ($\overline{T}$) are reported.
  %\edward{retrained from scratch - assessed performance when trained from scracth on the same data. Data efficiency. Molscribe is quite complex - transformer, we inject more expert knowledge. Graph constructor is not trained.}
  }
  \label{tab:training}
\end{table}

\subsection{Fine-grained model evaluation}
\begin{figure}[h]
  \centering
  %raw data for plot:
  %types = ("C", "single", "double","triple","O","N","S","F","Cl","P")
%count_means = {
%    'Ours fine-tuned': (0.712, 0.59, 0.759, 0.642, 0.871, 0.892, 0.846, 0.868, 0.863, 0.916),
%    'Decimer fine-tuned v2.2.0': (0.808, 0.806, 0.898, 0.731, 0.88, 0.858, 0.899, 0.728, 0.738, 0.719)),
%
%'ChemExpert([2],[1])':(0.84,0.83,0.94,0.81,0.93,0.9,0.93,0.81,0.77,0.75),}
   %\includegraphics[width=1\linewidth]{author-kit-CVPR2024-v2/figures/barchart_handdrawn_plus.png}
   %\includegraphics[width=0.85\linewidth,natwidth=800,natheight=800]{figures/polar_handdrawn_plus_1.png}
   \includegraphics[width=0.85\linewidth]{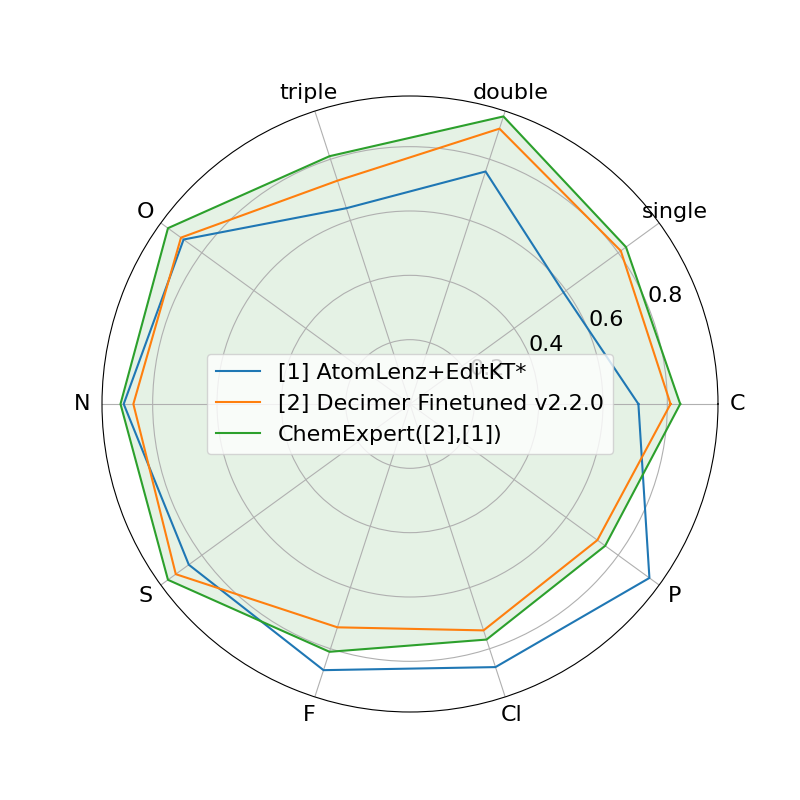}

   \caption{Count accuracies per type over images if type is present in image for hand-drawn test set. We observe errors of '\methodstepA+\methodstepB' and 'DECIMER fine-tuned' tend
to occur on different samples. Combining both approaches in \methodstepC~ improves performance.}
   \label{fig:perfplot2}
\end{figure}

%\edward{Todo: improve this paragraph}
Additionally, we conduct a detailed performance analysis of the most effective models from Table~\ref{tab:hand_drawn_combined}, presented in Figure~\ref{fig:perfplot2}. This figure showcases the count accuracies per atom or bond type. For each specific type, we identify images featuring that particular atom or bond type, then examine the predictions made by the methods on these images. Subsequently, we calculate the count accuracies for the predicted objects of the specific type within these images. For instance, when analyzing the 'triple bond' type, we select test images where at least one triple bond is depicted in the molecule and evaluate whether the method accurately predicts the correct number of triple bonds in the resulting molecular graph.

\begin{figure}[h]
  \centering
%types = ("C", "single", "double","triple")
%count_means = {
%    'Ours fine-tuned': (0.59, 0.576, 0.734, 0.574),
%    'Decimer fine-tuned v2.2.0': (0.568, 0.641, 0.742, 0.84),
%'ChemExpert([2],[1])':(0.58,0.65,0.75,0.85),
   %\includegraphics[width=1\linewidth]{author-kit-CVPR2024-v2/figures/barchart_chempix_plus.png}
   %\includegraphics[width=0.5\linewidth,natwidth=800,natheight=800]{figures/polar_chempix_plus_1.png}
   \includegraphics[width=0.5\linewidth]{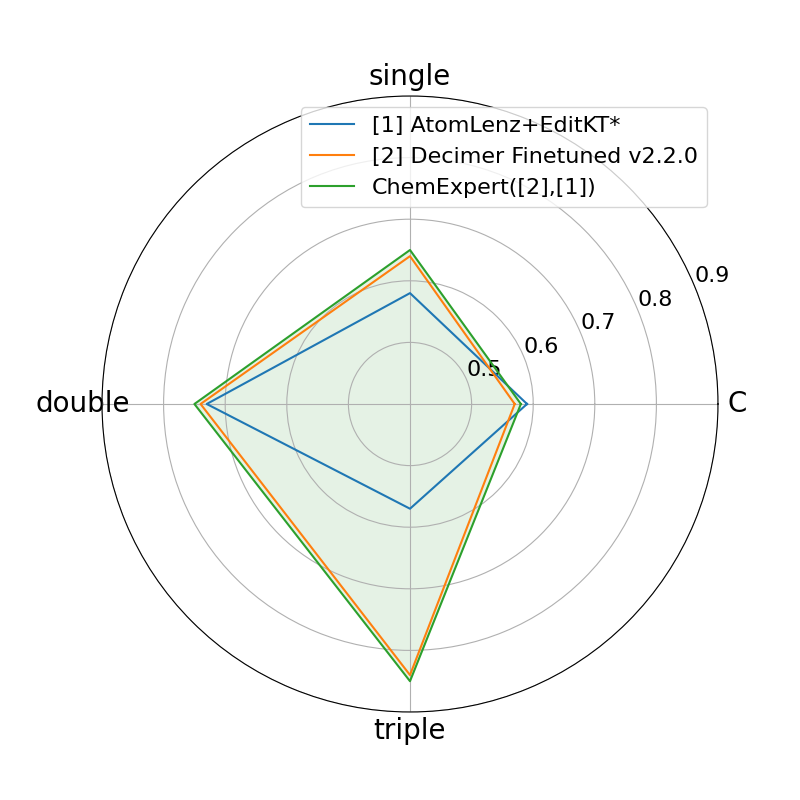}

   \caption{Count accuracies per type over images if type is present in image for ChemPix test set. We observe errors of '\methodstepA+\methodstepB' and 'DECIMER fine-tuned' tend
to occur on different samples. Combining both approaches in \methodstepC~ improves performance.}
   \label{fig:sm_perfplot4}
\end{figure}

The plot in Figure~\ref{fig:perfplot2} exhibits distinct patterns between `\methodstepA+\methodstepB~' and Decimer fine-tuned~\cite{decimerai}. For example `\methodstepA+\methodstepB~' performs better on images with Chlorine (Cl), Fluorine (F), and Phosphorus(P) compared to Decimer fine-tuned but worse on bonds. This variability may clarify why combining both predictions into \methodstepC~ leads to improved performance, as errors tend to occur on different samples, and the two approaches complement each other. The same analysis is performed for the ChemPix dataset in the SM section~\ref{sec:allresults} in Figure~\ref{fig:sm_perfplot4}.

\section{Conclusion}

This study has undertaken a comprehensive evaluation of various methods for chemical structure recognition, with a primary focus on the challenging domain of hand-drawn images. Our findings reveal insights into the strengths and limitations of existing tools and provide a compelling case for the efficacy of our approach. We showed that our method fares competitively despite a lower number of training samples, and resulted in state-of-the-art performance when combined with previous approaches. Our experiments highlighted our method's proficiency in precisely localizing atom-level entities, a feature notably lacking in many existing tools. Importantly, we showed that our architecture is remarkably more data-efficient than previous models

Despite these improvements in chemical structure recognition, reliably predicting the molecular structure from hand-drawn remains a challenge, and higher prediction performance would be required for a wide adoption of these tools. We hope that the release of our curated hand-drawn molecules images dataset, with detailed atom-level annotations, to the community will contribute to the development of more efficient and reliable tools.

\subsubsection*{Acknowledgments} %for camera ready only
AA, MO and YM are funded by (1) Research Council KU Leuven: Symbiosis 4 (C14/22/125), Symbiosis3 (C14/18/092); (2) Federated cloud-based Artificial Intelligence-driven platform for liquid biopsy analyses (C3/20/100); (3) CELSA - Active Learning (CELSA/21/019); (4) European Union's Horizon 2020 research and innovation programme under the Marie Skłodowska-Curie grant agreement No. 956832; (5) Flemish Government (FWO: SBO (S003422N), Elixir Belgium (I002819N), SB and Postdoctoral grants: S003422N, 1SB2721N, 1S98819N, 12Y5623N) and (6) VLAIO PM: Augmenting Therapeutic Effectiveness through Novel Analytics (HBC.2019.2528); (7) YM, AA, and MO are affiliated to Leuven.AI and received funding from the Flemish Government (AI Research Program). Computational resources and services used in this work were partly provided by the VSC (Flemish Supercomputer Center), funded by the Research Foundation - Flanders (FWO) and the Flemish Government – department EWI.

\clearpage
{
    \small
    \bibliographystyle{ieeenat_fullname}
    \bibliography{main}
}

\clearpage
\setcounter{page}{1}
\onecolumn
\maketitlesuppl
%\begin{multicols}{3}
%This document is supplementary material for the paper "Advancing Chemical Structure Recognition in Hand-Drawn Images by Atom-Level Entity Localization" paper id \#14628.
%\onecolumn
\section{Source code and datasets}\label{app:code}

\subsection{Source code}
The source code and basic instructions are available on
\url{https://github.com/molden/atomlenz}
%\url{https://anonymous.4open.science/r/atomlenz-B0C4}.

\subsection{Datasets}
Several datasets were used in this work and all are available.

\subsubsection{Hand-drawn images dataset}

The dataset introduced by \citet{handdrawndataset}, which consists of hand-drawn chemical depictions matched with their corresponding SMILES representations, is partitioned into 4,070 samples for training and validation purposes, along with an additional 1,018 samples for testing. These sets are referred to as the \emph{hand-drawn training set} and the \emph{hand-drawn test set} and available here:
\url{https://dx.doi.org/10.6084/m9.figshare.24599412}%\url{https://figshare.com/s/82a77a01906fea121fe1}

The \emph{hand-drawn training set} was then relabeled using \methodstepB~ to annotate
corresponding bounding box labels for 1417 images (see Experiments Section \ref{training_eff}). The format of the bounding box labels are further explained in Section~\ref{sec:synthetic_dataset}.
The dataset is available here:
\url{https://dx.doi.org/10.6084/m9.figshare.24599172}
%\url{https://figshare.com/s/a28899ea79dc75b30ba6}.

To streamline the process, the \emph{hand-drawn training set} is offered in different formats together with instructions to assist in training other baseline models (see Experiments Section \ref{training_eff}). When applicable, localization annotations are also included. The different datasets are available here:
\begin{itemize}
    \item DECIMER format:
    \url{https://dx.doi.org/10.6084/m9.figshare.24591252}%\url{https://figshare.com/s/9b168ecd7b4dbc3a9240}
    \item Img2Mol format    \url{https://dx.doi.org/10.6084/m9.figshare.24591381}%\url{https://figshare.com/s/10b4139e8cf44ff6d1b8}
    \item MolScribe format
    \url{https://dx.doi.org/10.6084/m9.figshare.24591300}%\url{https://figshare.com/s/1c5ea3206d7ce2a81a6f}
    \item ChemGrapher format
    \url{https://dx.doi.org/10.6084/m9.figshare.24591495}%\url{https://figshare.com/s/28bdb3df7c07da726e6d}
\end{itemize}

\subsubsection{Synthetically generated dataset}\label{sec:synthetic_dataset}
For the pretraining of the object detection models of \methodstepA , we generate images synthetically using RdKit~\cite{rdkit} and Indigo~\cite{pavlov2011indigo} paired with bounding boxes delineating all objects within, including atoms, bonds, charges, and stereocenters, similarly to what is used in other chemical structure recognition tools~\cite{oldenhof2020chemgrapher,qian2023molscribe}. Specifically, we collect approximately 214,000 chemical compounds in SMILES format from the ChEMBL~\cite{gaulton2017chembl} database. To enhance the method's resilience to stylistic variations, we introduce variability in elements such as fonts, font sizes, line widths, and the spacing between multiple bonds during image generation. Dataset is available in 2 parts: \begin{itemize}
    \item \textbf{part 1} atom and bond entity annotated images: \url{https://zenodo.org/records/10185264}
    \item \textbf{part 2} charge and stereocenter entity annotated images: \url{https://zenodo.org/records/10200185}
\end{itemize}

Example label file:

\begin{verbatim}
label,xmin,ymin,xmax,ymax
0,267,522,286,541
2,317,489,336,508
0,313,429,332,448
0,363,396,382,415
2,360,337,379,356
0,306,310,325,329
2,256,343,275,362
0,370,516,389,535
0,374,576,393,595
0,428,603,447,622
2,478,570,497,589
0,474,510,493,529
2,524,477,543,496
0,578,504,597,523
0,628,471,647,490
0,682,498,701,517
0,732,465,751,484
0,728,405,747,424
0,675,378,694,397
6,431,663,450,682
3,581,564,600,583
6,671,318,690,337
0,260,403,279,422
0,421,483,440,502
0,625,411,644,430
\end{verbatim}

Above an example csv label file is illustrated of the bounding box labels for one image. There are several fields in the csv file:

\begin{itemize}
    \item \textbf{label} field will annotate for every bounding box in the image what class the atom,bond,charge or stereo center the entity belongs to. For atom-type entities these are the different possible labels:
    \begin{verbatim}
        {0: 'C', 1: 'H', 2: 'N', 3: 'O', 4: 'S', 5: 'F', 6: 'Cl',
    7: 'Br', 8: 'I', 9: 'Se', 10: 'P', 11: 'B', 12: 'Si',
    13: '*', 14:'Te', 15:'Sn', 16: 'As', 17:'Al', 18:'Ge',
    19:'D', 20:'T'}
    \end{verbatim}
     For bond-type entities the different possible labels are:
     \begin{verbatim}
        {1: 'single', 2: 'double', 3: 'triple',
        4: 'aromatic', 5: 'wedged', 6: 'dashed'}
    \end{verbatim}
    For charge-type entities the different possible labels are:
    \begin{verbatim}
        {0: 0, 1: +1, 2: -1, 3: +2, 4: -2, 5: +3, 6: +4, 7: +5, 8: +6}
    \end{verbatim}
    Finally for stereocenters entities:
    \begin{verbatim}
        {0:'stereocenter'}
    \end{verbatim}
    \item \textbf{xmin,ymin:} coordinates of top left corner of the bounding box.
    \item \textbf{xmax,ymax:} coordinates of the bottom right corner of the bounding box.
\end{itemize}

\begin{figure*}
\begin{tcolorbox}[title={Illustrations datasets samples},standard jigsaw,opacityback=0,colbacktitle=white,coltitle=black,]
\begin{subfigure}{0.3\textwidth}
\centering
\includegraphics[scale=0.1]{figures/36.png}
\includegraphics[scale=0.1]{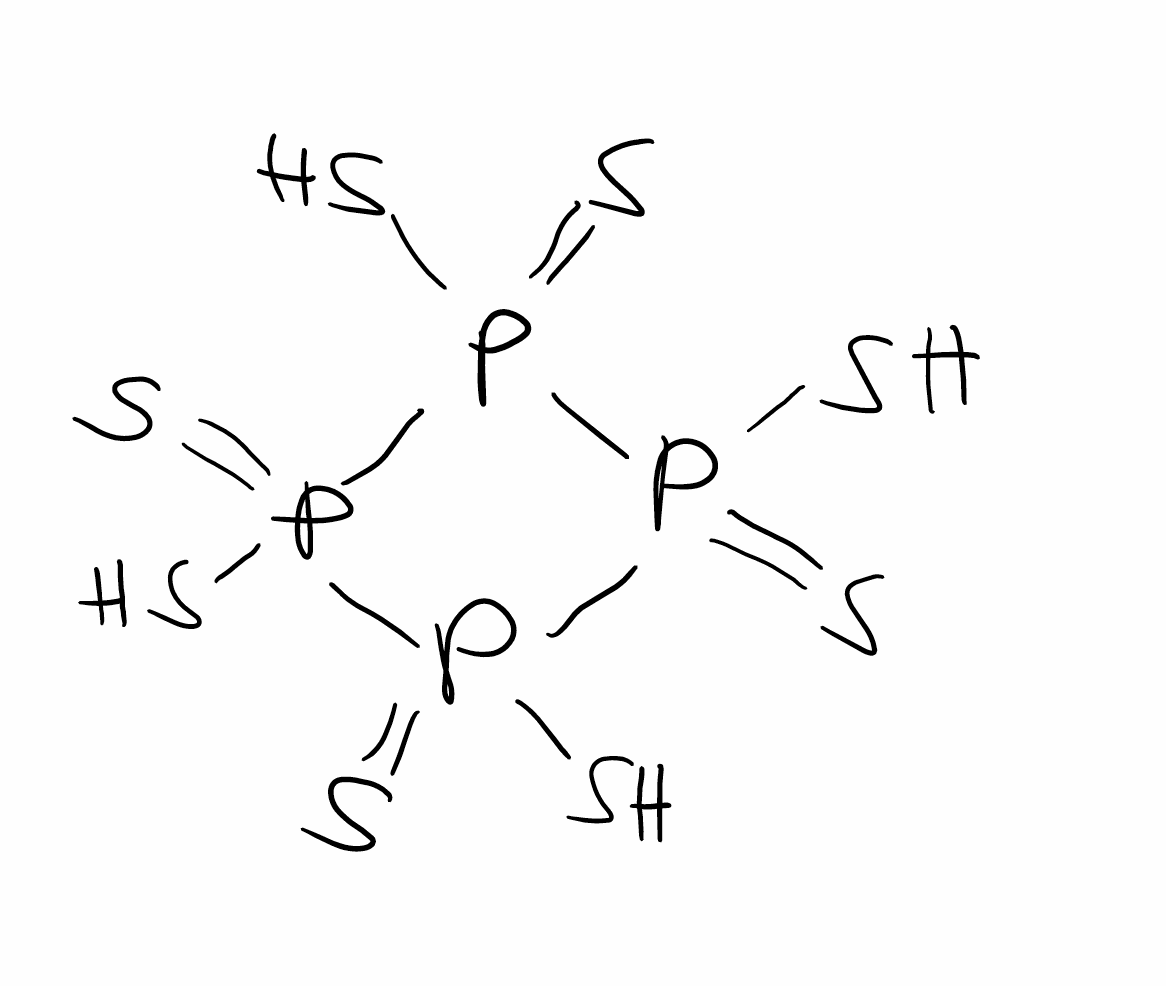}
\caption{Hand-drawn test set}
%\label{fig:style_sub1}
\end{subfigure}
\begin{subfigure}{0.3\textwidth}
\centering
\includegraphics[scale=0.4]{figures/ring_chempix.png}
\includegraphics[scale=0.4]{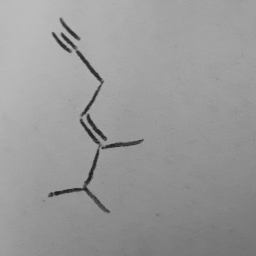}
\caption{Chempix test set.}
%\label{fig:style_sub3}
\end{subfigure}
\begin{subfigure}{0.3\textwidth}
\centering
\includegraphics[scale=0.4]{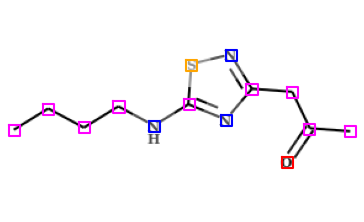}
\includegraphics[scale=0.38]{figures/example_image_cut.png}
\caption{Atom localization test set, labeled with bounding boxes, here illustrated with different color for every atom type.}
%\label{fig:style_sub4}
\end{subfigure}
\end{tcolorbox}
\caption{Different example samples for the different datasets used in experiments.}
\label{fig:example_samples_long}
\end{figure*}

Examples of samples from the synthetically generated training set are illustrated in Figure~\ref{fig:example_train} together with the drawn bounding box labels for the different object types.
Also some extra examples of samples of all test sets described in Section~\ref{sec:datasets} are illustrated in Figure~\ref{fig:example_samples_long}.
\clearpage
%\end{multicols}

\section{Illustrations of isomers}
\label{sec:isomer}
In all imperfect representation levels there are compounds that cannot be distinguished. These undistinguishable groups correspond to the concept of isomerism in chemistry. In the level of molecular formula where only the count of different atoms are given, these equivalent compounds called constitutional isomers. Compound graphs with identical adjacency matrix but different spatial organization are called stereoisomers. In the following we give examples to help clarify these concepts.

\subsection{Constitutional isomerism}
Constitutional isomerism is a quite simple concept. It is clear that if we specify the number of atoms for all type, multiple possible compound graphs can be built. There are valence constraints of course, for example a $C_nH_{2n+2}$ compound cannot contain double or triple bonds. However, there is still a large variety of graphs that can be realized.

\begin{figure*}
\centering
\begin{subfigure}{0.45 \textwidth}
  \centering
  \includegraphics[scale=0.5]{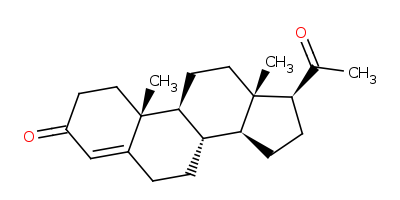}
   \caption{progesterone (chemical formula C$_{21}$H$_{30}$O$_2$).}
   \label{fig:sup:progesterone}
\end{subfigure}
\hfill
\begin{subfigure}{0.45 \textwidth}
  \centering
  \includegraphics[scale=0.5]{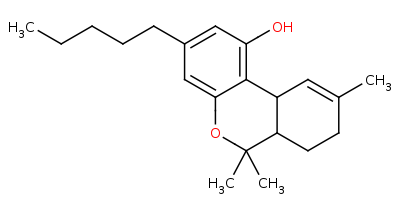}
   \caption{$\Delta$9-tetrahydrocannabinol (chemical formula C$_{21}$H$_{30}$O$_2$).}
   \label{fig:sup:THC}
\end{subfigure}
\caption{Constitutional isomerism between two unrelated compounds.}
\label{sup:fig:progthc}
\end{figure*}

\begin{figure*}
\begin{subfigure}{0.45 \textwidth}
  \centering
\includegraphics[scale=0.6]{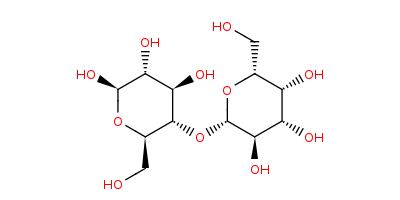}
   \caption{Depiction of lactose (chemical formula C$_{12}$H$_{22}$O$_{11}$).}
   \label{fig:sup:lactose}
\end{subfigure}
\begin{subfigure}{0.45 \textwidth}
  \centering
   \includegraphics[scale=0.6]{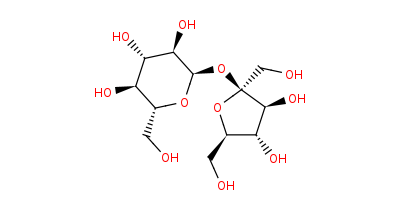}
   \caption{Depiction of sucrose (chemical formula C$_{12}$H$_{22}$O$_{11}$).}
   \label{fig:sup:sucrose}
\end{subfigure}
   \caption{Constitutional isomerism between two compounds from the same family (disacharides).}
\label{sup:fig:lactsuc}
\end{figure*}

The case of sucrose and lactose (see Figure \ref{sup:fig:lactsuc}) is easy to follow as the galactose and fructose unit only different in the position of the ring closure. A more accidental case is progesterone and THC depicted on Figure \ref{sup:fig:progthc}. This starkly illustrates the pitfalls of using the chemical formula as a representation. The effects of these two compounds are clearly unrelated.

% -----

\subsection{Stereoisomerism}

If we only take into account the atom and bond adjacency relations we have some relevant degree of freedom undescribed. An often used example is our hands. While all bones have the same adjacency in both of our hands, we cannot rotate the two such that they are identical: they are mirror images.

Note two important details. Firstly, we do not care about exact positions of atoms in 3D space when the molecule is flexible, similarly as we do not distinguish a hand with closed  or opened fingers, but differentiating between the left and right hand.
Secondly, the spatial organization has nothing to do with the placement of the atoms on the 2D depiction plane, these positions are arbitrary.

\begin{figure*}
\centering
\begin{subfigure}{0.45 \textwidth}

   \includegraphics[scale=0.5]{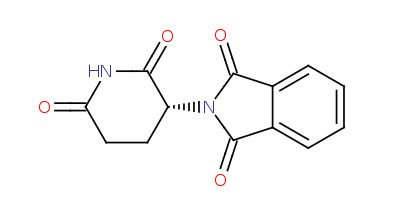}

   \caption{Depiction of (R)-Thalidomide, a compound with sedative effect.}
   \label{fig:R_thalidomide}
\end{subfigure}
\hfill
\begin{subfigure}{0.45 \textwidth}
  \centering
   \includegraphics[scale=0.5]{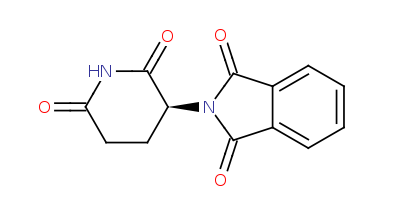}

   \caption{Depiction of (S)-Thalidomide, a CRBN targeting compound with teratogenic effect}
   \label{fig:S_thalidomide}
\end{subfigure}
\caption{Illustration of stereochemisty and its depiction: wedge bonds.}
\label{sup:fig:thalidomide}
\end{figure*}

To enhance our representation, new labels need to be introduced: wedge bonds and/or stereocenters. For example see Figure \ref{sup:fig:thalidomide}.
The filled wedge bond indicates that the atom or group at the thick end pointing out of the plane of the drawing, while the dashed wedge bond indicates that the group is under that plane. The depicted compounds are mirror images of each other, however, the difference in the biological effect can be dramatic (in the case of thalidomide the picture is more complicated, as the two form can interconvert in the body, but for didactic pourposes let us assume this is not the case).
Note that if the left ring would be symmetric, for example by connecting the nitrogen to the neighboring carbon, the two compound would be identical. A simple 180 degree rotation around the long axis of the compound would show this.
Stereoisomerism necessitates the presence of an atom lacking symmetric surroundings. This unique atom, such as the carbon at the wedge bond in this scenario, is referred to as a stereo center.

\begin{figure*}
\begin{subfigure}{0.4 \textwidth}
  %\centering

   %\includegraphics[scale=0.5,natwidth=400,natheight=200]{figures/D-Threose.png}
   \includegraphics[scale=0.5]{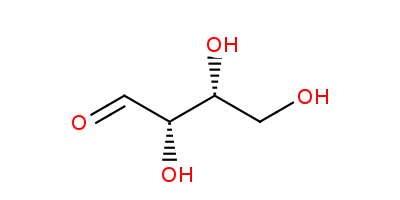}

   \caption{(D)-Threose}
   \label{fig:D_threose}
\end{subfigure}
\hfill
\begin{subfigure}{0.4 \textwidth}
%  \centering

   %\includegraphics[scale=0.5,natwidth=400,natheight=200]{figures/L-Threose.png}
   \includegraphics[scale=0.5]{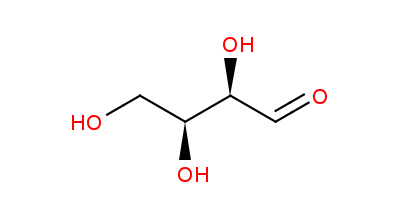}
   \caption{(L)-Threose}
   \label{fig:L_threose}
\end{subfigure}
\hfill
\begin{subfigure}{0.4 \textwidth}
 % \centering

   %\includegraphics[scale=0.5,natwidth=400,natheight=200]{figures/D-Erythrose.png}
   \includegraphics[scale=0.5]{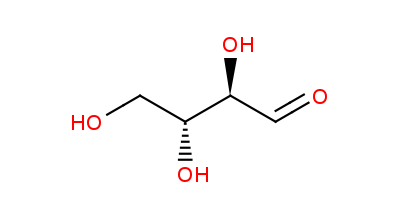}
   \caption{(D)-Erythrose}
   \label{fig:D_erythreose}
\end{subfigure}
\hfill
\begin{subfigure}{0.4 \textwidth}
  \centering
   \includegraphics[scale=0.5]{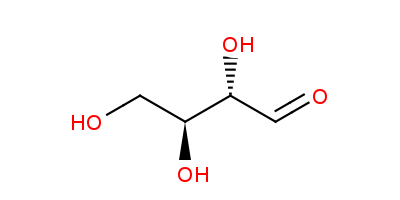}
   \caption{(L)-Erythrose}
   \label{fig:L_erythreose}
\end{subfigure}
\caption {Illustration of stereoisomeric relations: enantiomers and diastereomers. All four compound differs only in the orientation of the two OH grops. While the compounds on the right can be transformed to the compound on the left by mirroring (they are enantiomers in pairs) the compounds in top of each other cannot (they are diastereomers)}
\label{sup:fig:stereo}
\end{figure*}

Stereoisomers are not always mirror images of each other. If there are $n$ stereocenters  in a molecule (see Figure \ref{sup:fig:stereo}) there are $2^n$ stereoisomers, forming pairs of mirror images (called enantiomers). The non-mirror image pairs are called diastereomers.

\section{Details of graph algorithm subroutines}
\label{sec:subroutines}
This section aims to provide in-depth insights into the subroutines utilized within the molecular graph constructor as introduced in Algorithm~\ref{alg:alg-1}.

\begin{itemize}
    \item The first subroutine used in the molecular graph constructor is $\mathrm{filterAtoms}(\mathbf{O}^a)$. This subroutine is implemented inside the function \verb!iou_filter_bboxes! in the file \verb!utils_graph.py! .

    The function goes over all overlapping bounding boxes of atoms with IoU higher than 0.5. For every group of overlapping bounding boxes the function will keep the bounding box with the highest score.

    \item $\mathrm{checkCharges}(\mathbf{O}^c, \mathbf{o}^a)$ is responsible for determining which atom objects should carry a charge and is implemented in \verb|predict_smiles.py| from line \verb|95| until \verb|99|:
    \begin{verbatim}
95 charge_atoms = np.ones(len(filtered_bboxes))
96 for index,box_atom in enumerate(filtered_bboxes):
97    for box_charge,label_charge in zip(filtered_ch_boxes,filtered_ch_labels):
98        if bb_box_intersects(box_atom,box_charge) == 1:
99            charge_atoms[index]=label_charge
    \end{verbatim}
    \item $\mathrm{checkStereoChem}(\mathbf{O}^s, \mathbf{o}_{c}^a)$  is applied to identify atoms functioning as stereocenters and is implemented in \verb|predict_smiles.py| from line \verb|141| until \verb|151|:
\begin{verbatim}
141 stereo_bonds = np.where(mol_graph>4, True, False)
142 if np.any(stereo_bonds):
143       stereo_boxes = stereo_preds[image_idx]['boxes'][0]
144       stereo_labels= stereo_preds[image_idx]['preds'][0]
145       for stereo_box in stereo_boxes:
146         result=[]
147         for atom_box in filtered_bboxes:
148            result.append(bb_box_intersects(atom_box,stereo_box))
149         indices = [i for i, x in enumerate(result) if x == 1]
150         if len(indices) == 1:
151            stereo_atoms[indices[0]]=1
\end{verbatim}
    \item $\mathrm{checkEdge}(V, \mathbf{o}^b)$ evaluates which vertices (atoms) overlap with the bonds and is implmented in \verb|predict_smiles.py| from line \verb|109| until \verb|118|:
\begin{verbatim}
109 result = []
110 limit = 0
111
112 while result.count(1) < 2 and limit < 80:
113      result=[]
114      bigger_bond_box = [bond_box[0]-limit,
         bond_box[1]-limit,bond_box[2]+limit,bond_box[3]+limit]
115      for atom_box in filtered_bboxes:
116         result.append(bb_box_intersects(atom_box,bigger_bond_box))
117     limit+=5
118 indices = [i for i, x in enumerate(result) if x == 1]
\end{verbatim}
\item $\mathrm{filterCands}(\mathrm{candAtoms})$ will select the two most probable atoms to form a bond when more than 2 candidate atoms appear. This step is implemented in \verb|dist_filter_bboxes(cand_bboxes)| in file \verb|utils.graph.py|.

\item Finally the validation step is performed by performing several iterations this code:
\begin{verbatim}
    mol =  Chem.MolFromMolFile('molfile',sanitize=False)
    problematic = 0
    try:
       problems = Chem.DetectChemistryProblems(mol)
       if len(problems) > 0:
          mol = solve_mol_problems(mol,problems)
\end{verbatim}
Where \verb|solve_mol_problems| is implemented in file \verb|utils_graph.py|.

\end{itemize}

%\section{Stereochemistry}
%\label{sec:stereo chemistry}
\section{Illustrations of types of atom-level entities}
\label{sec:examples_detection}
\begin{figure*}
\begin{tcolorbox}[title={Illustrations of types of atom-level entities},standard jigsaw,opacityback=0,colbacktitle=white,coltitle=black,]
\begin{subfigure}{0.3\textwidth}
\centering
\includegraphics[scale=0.5]{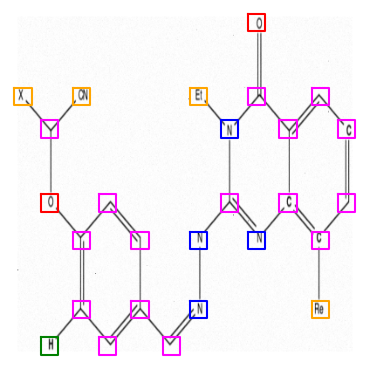}
\caption{atom type entities}
\label{fig:atom_sub1}
\end{subfigure}
\begin{subfigure}{0.3\textwidth}
\centering
\includegraphics[scale=0.5]{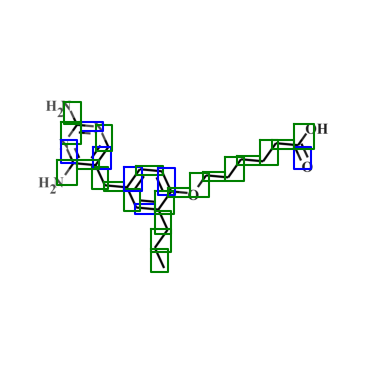}
\caption{bond type entities}
\label{fig:bond_sub3}
\end{subfigure}
\begin{subfigure}{0.3\textwidth}
\centering
\includegraphics[scale=0.55]{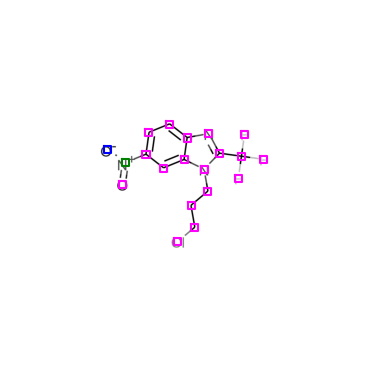}
\caption{charge type entities}
\label{fig:charge_sub4}
\end{subfigure}
\begin{subfigure}{0.3\textwidth}
\centering
\includegraphics[scale=0.7]{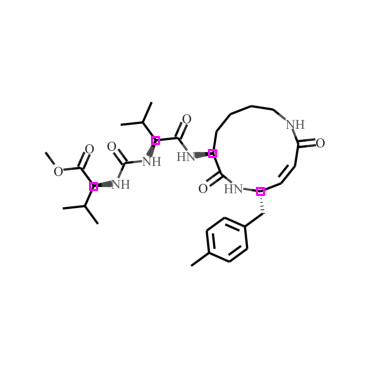}
\caption{stereocenter type entities}
\label{fig:stereo_sub4}
\end{subfigure}
\end{tcolorbox}
\caption{Different illustrations of types of atom-level entities}
\label{fig:example_train}
\end{figure*}

Examples of samples from the synthetically generated training set are illustrated in Figure~\ref{fig:example_train} together with the drawn bounding box labels for the different atom-level entity types: atoms, bonds, charges and stereocenters.

\section{All results}
In our experiments we assess the molecular structure prediction performance using accuracy and Tanimoto similarity, a widely used metric for quantifying molecular similarity, to assess the resemblance between the model’s predictions and the actual molecular graphs. Tanimoto similarity values range from 0 to 1, with higher values indicating greater similarity. A Tanimoto similarity of 1 indicates that the structural descriptors are identical or that they are matching `on-bits' in a binary fingerprint. The binary fingerprint employed to measure the Tanimoto similarity is the Extended-connectivity fingerprint~\cite{ecfp} with radius 3 (ECFP6) and fingerprint length of 2048.
Crafted with precision to capture essential molecular features relevant to molecular activity, ECFPs (Extended-Connectivity Fingerprints)~\cite{ecfp} are generated through a customized adaptation of the Morgan~\cite{morgan} algorithm. This involves systematically traversing each atom in the molecule to extract all possible paths within a specified radius. Following this, every unique path undergoes hashing into a numerical value within a predetermined bit range. It is worth noting that the encoded fragment size expands proportionally with an increased radius.

Our Tables~\ref{tab:allresults1} and \ref{tab:allresults2} report both the accuracy, computed by counting the instances where the predicted structures have identical structural ECFP6 descriptors (denoted by a Tanimoto similarity of 1) and the average Tanimoto similarity. As an additional metric, we include the accuracy when assessing whether the predicted resulting SMILES exactly match the true SMILES.

Lastly, we conduct supplementary experiments utilizing \methodstepC~ on both the Chempix and hand-drawn test sets, while altering the sequence of chemical structure tools. In both datasets, we note that the combined utilization of \methodstepA+\methodstepB~ and DECIMER fine-tuned within \methodstepC~ yields the best performance. Nevertheless, the arrangement of tools within \methodstepC~ slightly alters the performance, depending on the test set and the specific performance metric, as demonstrated in Table~\ref{tab:allresults2}.

\label{sec:allresults}

\begin{table*}
  \centering
  \begin{tabular}{@{}lccc@{}}
    \toprule
    Method & Acc. (exact match) & Acc.($T=1$) & $\overline{T}$ \\
    \midrule
    DECIMER (v2.2.0) \cite{rajan2020decimer}& 0.281 & 0.295 & 0.451 \\
    DECIMER fine-tuned(v2.2.0) \cite{decimerai}& 0.567 & 0.622 & 0.727 \\
    Img2Mol \cite{clevert2021img2mol}& 0.047 & 0.084 & 0.275 \\
    MolScribe \cite{qian2023molscribe}& 0.094 & 0.102 &  0.288 \\
    ChemGrapher \cite{oldenhof2020chemgrapher}&0.002 & 0.002 &  0.065 \\
    OSRA \cite{filippov2009optical}& 0.006 & 0.006 &  0.065 \\
    \midrule
    \methodstepA & 0.008 & 0.009 &  0.087 \\
    \methodstepA+\methodstepB & 0.279 & 0.338 & 0.484 \\
    \methodstepC(\methodstepA+\methodstepB,\cite{decimerai}) & 0.416 & 0.417 & 0.572 \\
    \methodstepC(\cite{decimerai},\cite{rajan2020decimer}) & 0.571 & 0.626 & 0738\\
     \methodstepC(\cite{decimerai},\methodstepA+\methodstepB) & \textbf{0.579} &  \textbf{0.635} &  \textbf{0.749}\\
    \bottomrule
  \end{tabular}
  \caption{Benchmark results on target domain (hand-drawn images test set). Both the accuracy, computed by counting the instances where the predicted structures have identical structural ECFP6 descriptors (denoted by a Tanimoto ($T$) similarity of 1) and the average Tanimoto similarity ($\overline{T}$) are reported. As an additional metric, we include the accuracy when assessing whether the predicted resulting SMILES exactly match the true SMILES.}
  \label{tab:allresults1}
\end{table*}

\begin{table*}
  \centering
  \begin{tabular}{@{}lccc@{}}
    \toprule
    Method & Acc. (exact match) & Acc.($T=1$) & $\overline{T}$ \\
    \midrule
    DECIMER (v2.2.0) \cite{rajan2020decimer}& 0.036 & 0.05 & 0.1 \\
    DECIMER fine-tuned (v2.2.0) \cite{decimerai}& 0.482 & 0.508 & 0.643 \\
    Img2Mol \cite{clevert2021img2mol}& 0.015 & 0.015 & 0.084 \\
    MolScribe \cite{qian2023molscribe}& 0.228 & 0.269 &  0.417 \\
    ChemGrapher \cite{oldenhof2020chemgrapher}& 0.151 & 0.187 &  0.286 \\
    OSRA\cite{filippov2009optical} & 0.044 & 0.047 &  0.071 \\
    \midrule
    \methodstepA & 0.026 & 0.054 &  0.064 \\
    \methodstepA+\methodstepB & 0.4  & 0.484 & 0.605 \\
    \methodstepC(\methodstepA+\methodstepB,\cite{qian2023molscribe}) & 0.412 & 0.5 & 0.619 \\
    \methodstepC(\methodstepA+\methodstepB,\cite{decimerai}) & 0.441 & \textbf{0.529} & 0.65 \\
    \methodstepC(\cite{decimerai},\methodstepA+\methodstepB) & \textbf{0.487} & 0.518 & \textbf{0.655} \\
    \bottomrule
  \end{tabular}
  \caption{Benchmark results on out of domain ChemPix test set. Both the accuracy, computed by counting the instances where the predicted structures have identical structural ECFP6 descriptors (denoted by a Tanimoto ($T$) similarity of 1) and the average Tanimoto similarity ($\overline{T}$) are reported. As an additional metric, we include the accuracy when assessing whether the predicted resulting SMILES exactly match the true SMILES.}
  \label{tab:allresults2}
\end{table*}

\begin{table*}
  \centering
  \begin{tabular}{@{}lcccc@{}}
    \toprule
    Method & Acc.($T=1$) (test set) & $\overline{T}$   (test set) & Acc.($T=1$) (train set) & $\overline{T}$ (train set) \\
    \midrule
    DECIMER (v2.2.0) \cite{rajan2020decimer}& 0.001 & 0.039  & 0.099 & 0.142\\
    Img2Mol \cite{clevert2021img2mol}& 0.0 & 0.0867 & 0.237 & 0.388  \\
    MolScribe \cite{qian2023molscribe}& 0.013 &  0.0865 & 0.234 & 0.275 \\
    ChemGrapher \cite{oldenhof2020chemgrapher}& 0.004 & 0.067 &  0.007 & 0.073\\
%    OSRA\cite{filippov2009optical} & - &  - \\
    \midrule
    \methodstepA & \textbf{0.338} &  \textbf{0.484} & \textbf{0.383} & \textbf{0.522}  \\
    \bottomrule
  \end{tabular}
  \caption{All methods are retrained from scratch on same training dataset (4070 samples of hand-drawn images) to asses data efficiency. Benchmark results on both hand-drawn images train and test set. Both the accuracy, computed by counting the instances where the predicted structures have identical structural ECFP6 descriptors (denoted by a Tanimoto ($T$) similarity of 1) and the average Tanimoto similarity ($\overline{T}$) are reported.}
  \label{tab:alltraining}
\end{table*}

%
%Having the supplementary compiled together with the main paper means that:
%
%\begin{itemize}
%\item The supplementary can back-reference sections of the main paper, for example, we can refer to \cref{sec:intro};
%\item The main paper can forward reference sub-sections within the supplementary explicitly (e.g. referring to a particular experiment);
%\item When submitted to arXiv, the supplementary will already included at the end of the paper.
%\end{itemize}
%
%To split the supplementary pages from the main paper, you can use \href{https://support.apple.com/en-ca/guide/preview/prvw11793/mac#:~:text=Delete%20a%20page%20from%20a,or%20choose%20Edit%20%3E%20Delete).}{Preview (on macOS)}, \href{https://www.adobe.com/acrobat/how-to/delete-pages-from-pdf.html#:~:text=Choose%20%E2%80%9CTools%E2%80%9D%20%3E%20%E2%80%9COrganize,or%20pages%20from%20the%20file.}{Adobe Acrobat} (on all OSs), as well as \href{https://superuser.com/questions/517986/is-it-possible-to-delete-some-pages-of-a-pdf-document}{command line tools}.


\end{document}